\documentclass[letterpaper, 10 pt, conference]{ieeeconf}

\IEEEoverridecommandlockouts 

\makeatletter
\let\NAT@parse\undefined
\makeatother
\usepackage[numbers]{natbib}

\usepackage{amsmath} 
\usepackage{amssymb} 
\usepackage{bm}
\usepackage{xcolor}
\usepackage{graphicx}
\usepackage{mathtools}
\usepackage{hyperref}
\usepackage{cleveref}
\usepackage{siunitx}
\usepackage{tikz}
\usetikzlibrary{shapes}
\usepackage{multirow}
\usepackage{makecell}
\usepackage{cuted}
\usepackage{textcomp}
\usepackage{stfloats}
\usepackage{url}
\usepackage{verbatim}
\usepackage{cite}
\usepackage{gensymb}

\usepackage{subcaption}
\usepackage{tablefootnote}
\Crefformat{figure}{#2Fig.~#1#3}
\Crefmultiformat{figure}{Figs.~#2#1#3}{ and~#2#1#3}{, #2#1#3}{ and~#2#1#3}
\Crefformat{equation}{#2Eq.~#1#3}

\usepackage{xspace}
\makeatletter
\DeclareRobustCommand\onedot{\futurelet\@let@token\@onedot}
\def\@onedot{\ifx\@let@token.\else.\null\fi\xspace}
\makeatother

\newcommand{\algo}[1]{\textsc{#1}}
\newcommand{\method}{\algo{Rambo}\xspace}

\newcommand{\norm}[1]{\left\lVert#1\right\rVert}

\def\thickhline{\noalign{\hrule height.8pt}}

\definecolor{lightblue}{HTML}{9BB7D4} 
\definecolor{light2blue}{HTML}{92A8D1} 
\definecolor{aqua}{HTML}{7BC4C4} 
\definecolor{turquoise}{HTML}{53B0AE} 
\definecolor{iris}{HTML}{5A5B9F} 

\definecolor{darkyellow}{HTML}{F0C05A} 
\definecolor{yellow}{HTML}{F8D948} 

\definecolor{pink}{HTML}{D94F70} 
\definecolor{lightpink}{HTML}{F7CAC9} 
\definecolor{warmpink}{HTML}{FF6F61} 
\definecolor{darkpink}{HTML}{C74375} 
\definecolor{magenta}{HTML}{BB2649} 

\definecolor{orange}{HTML}{E2583E} 
\definecolor{lightorange}{HTML}{FEBE98} 
\definecolor{orangered}{HTML}{DD4124} 

\definecolor{green}{HTML}{009473} 
\definecolor{grassgreen}{HTML}{88B04B} 

\definecolor{redbrown}{HTML}{955251} 
\definecolor{darkred}{HTML}{9B1B30} 

\definecolor{purple}{HTML}{6968AC} 
\definecolor{darkpurple}{HTML}{5F4B8B} 
\definecolor{lightpurple}{HTML}{B163A3} 

\definecolor{grey}{HTML}{959A9C} 
\definecolor{beige}{HTML}{DECDBE} 

\hypersetup{
    colorlinks=true, 
    linkcolor=black, 
    citecolor=darkred, 
    filecolor=blue, 
    urlcolor=grey,
}
\usepackage{fancyhdr}
\fancypagestyle{firstpagefooter}{
  \fancyfoot[C]{%
    \footnotesize \copyright~2025 IEEE. Personal use of this material is permitted. Permission from IEEE must be obtained for all other uses, in any current or future media, including reprinting/republishing this material for advertising or promotional purposes, creating new collective works, for resale or redistribution to servers or lists, or reuse of any copyrighted component of this work in other works.
  }
}
\title{\LARGE \bf \method: RL-Augmented Model-Based Whole-Body Control \\ for Loco-Manipulation }

\author{Jin~Cheng$^{\alpha}$, Dongho~Kang$^{\alpha}$, Gabriele~Fadini$^{\alpha}$, Guanya~Shi$^{\beta}$, and Stelian~Coros$^{\alpha}$%
\thanks{$^{\alpha}$The authors are with Computational Robotics Lab in the Department of Computer Science, ETH Zurich, Zurich, Switzerland.
        {\tt\footnotesize \{jicheng, kangd, gfadini, scoros\}@ethz.ch}}%
\thanks{$^{\beta} $Guanya Shi is with the Robotics Institute and the School of Computer Science at Carnegie Mellon University, Pittsburgh, USA.
        {\tt\footnotesize guanyas@andrew.cmu.edu}}%
\thanks{Project website: \url{https://jin-cheng.me/rambo.github.io/}}%
\thanks{Supplementary video: \url{https://youtu.be/VdZxhLNG6wQ}}%
}

\begin{document}
    \maketitle

    \thispagestyle{firstpagefooter}
    \begin{abstract}
        Loco-manipulation, physical interaction of various
        objects that is concurrently coordinated with locomotion, remains a major challenge for legged robots due to the need for
        both precise end-effector control and robustness to unmodeled dynamics.
        While model-based controllers provide
        precise planning via online optimization, they are limited by model inaccuracies. In contrast, learning-based methods offer robustness, but they struggle
        with precise modulation of interaction forces. We introduce \method, a hybrid framework that integrates model-based
        whole-body control within a
        feedback policy trained with reinforcement learning. The model-based module
        generates feedforward torques by solving a quadratic program, while the
        policy provides feedback corrective terms to enhance robustness. We validate our framework on a quadruped robot across a diverse
        set of real-world loco-manipulation tasks, such as pushing a shopping cart,
        balancing a plate, and holding soft objects, in both quadrupedal and bipedal walking.
        Our experiments demonstrate that \method enables precise manipulation capabilities
        while achieving robust and dynamic locomotion.
    \end{abstract}
    
    
    \section{Introduction}
    \label{sec:introduction}
    Modern legged robots have demonstrated impressive mobility over a wide range
    of terrains~\citep{lee2020learning, miki2022learning, hoeller2024anymal,radosavovic2024real}.
    To expand their capabilities beyond conventional locomotion tasks, there is
    growing interest in loco-manipulation, which enables these machines to actively
    interact with and manipulate their surroundings. However, whole-body loco-manipulation
    remains a challenging task for these systems, as it requires coordinated control of both the base and end-effector movements to achieve precise and robust behaviors, which often pose conflicting objectives~\citep{gu2025humanoid}.

    Being robust against unmodeled effects and unexpected interactions, reinforcement learning (RL)-based
    controllers have achieved impressive results in various pedipulation and manipulation
    tasks~\citep{fu2023deep, ji2023dribblebot, arm2024pedipulate}. 
    However,
    training agents for loco-manipulation via RL
    still remains a challenging task.
    Often working in joint position space, learned policies tend to produce excessively large targets, which indirectly govern the resulting interaction forces~\citep{lyu2024rl2ac, ha2024learning}. In practice, this strategy prioritizes robustness at the expense of precision~\citep{cheng2023legs, zhang2024wococo}.
    Moreover, loco-manipulation tasks typically require exploration in a large joint space, necessitating motion prior, reward shaping, or other exploration strategies~\citep{jeon2023learning, schwarke2023curiosity, ha2024umi}.

    Model-based control methods, on the other hand, have proven highly effective
    for contact planning and handling interactions with objects in whole-body loco-manipulation
    tasks~\citep{bellicoso2019alma, sleiman2021unified, sleiman2023versatile}. 
    By explicitly taking contact forces into account, these approaches enable
    precise control and optimization of torque-level commands~\citep{di2018dynamic, kuindersma2016optimization, dai2014whole}.
    However, their performance heavily depends on how system dynamics is modeled, and on careful parameter identification on real hardware~\citep{wensing2023optimization}.
    In addition, model predictive control methods require a trade-off between model complexity, solution accuracy and planning horizon~\citep{kalakrishnan2011learning, grandia2019feedback}.

    The ultimate goal of this work is to equip the legged controllers with the capability
    to perform robust, precise, and efficient whole-body loco-manipulation. We aim to combine the strengths of model-based and learning-based approaches to achieve effective torque-level control while remaining robust against
    unmodeled effects and disturbances.

    \begin{figure}[t!]
        \vspace{0.2cm}
        \centering
        \includegraphics[width=\linewidth]{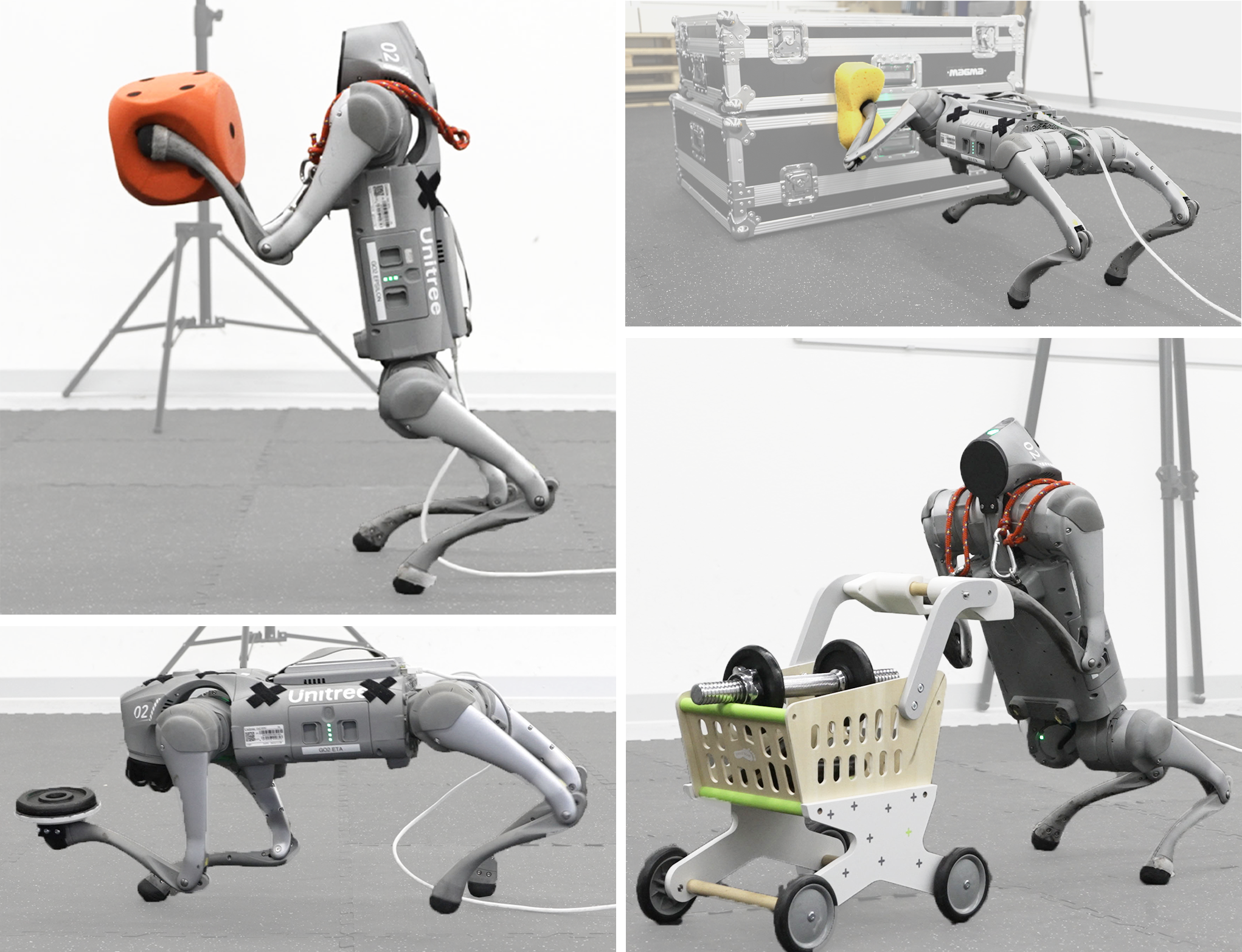}
        \caption{Various whole-body loco-manipulation tasks enabled by \method
        on Unitree Go2~\citep{go2} in both quadrupedal and bipedal modes.}
        \label{fig:teaser}
        \vspace{-0.6cm}
    \end{figure}

   To this end, we propose \method---\textcolor{darkred}{R}L-\textcolor{darkred}{A}ugmented
    \textcolor{darkred}{M}odel-\textcolor{darkred}{B}ased Wh\textcolor{darkred}{O}le-body
    Control---a hybrid control framework for whole-body loco-manipulation tasks
    on legged systems. Our method generates feedforward torques by optimizing
    end-effector contact forces through a model-based whole-body controller, formulated as a quadratic program (QP), while
    ensuring robustness with an RL policy that compensates for modeling
    errors through its corrective actions.

    We demonstrate the effectiveness of our method on a range of loco-manipulation
    tasks, including pushing a shopping cart, balancing a plate, and holding soft objects—spanning
    both quadrupedal and bipedal dynamic walking on a quadruped platform.
    Through extensive evaluations in simulated and real-world scenarios, \method demonstrates a high degree of precision in tracking end-effector
    targets while remaining robust in typical locomotion tasks.

    \section{Related Work}
    \label{sec:related_work}

    In this section, we review two major groups of the state-of-the-art control approaches
    for whole-body loco-manipulation applications: model-based and learning-based
    methods.

    \subsection{Model-based Methods for Loco-manipulation}
    \label{subsec:mblm}

    Model-based methods generate control signals by solving optimal control problems
    using first-principles dynamics models that describe the relationship between
    system states and control inputs~\citep{wensing2023optimization}. By explicitly
    accounting for interaction forces, these methods can effectively optimize
    motion while incorporating the dynamic effects from contacts into the
    control loop~\citep{kuindersma2016optimization, tassa2012synthesis}. When
    applied to loco-manipulation tasks, model-based approaches offer precise control
    over the forces exerted on objects and produce effective torque-level commands
    for each joint~\citep{sleiman2021unified, rigo2023contact}. Among these,
    Model Predictive Control (MPC) is especially popular for legged systems, as it
    provides a robust feedback mechanism and enables emergent behaviors by forecasting
    the impact of control inputs over a time horizon~\citep{sleiman2023versatile, khazoom2024tailoring}.
    However, the applicability of model-based methods can be limited in complex scenarios, such
    as locomotion over uneven terrain or manipulation of objects with involved
    inertial properties, where accurate knowledge of the environment is
    difficult to obtain and the required modeling and computational effort
    becomes excessively large~\citep{sleiman2023versatile, grandia2023perceptive}.

    State-of-the-art MPC methods for legged robots often resort to hierarchical frameworks where a middle-layer whole-body controller (WBC) is formulated to translate the high-level plans to torque-level commands through optimization in an instantaneous manner~\citep{grandia2023perceptive, jenelten2022tamols}.
    \method aims to preserve the benefits of torque-level control generated by model-based methods while overcoming their limitations.
    As such, our framework retains a computationally
    efficient whole-body control module~\citep{ding2019real} as a middle layer. 
    The WBC computes feedforward torques by explicitly optimizing reaction forces, and it is augmented with a policy trained via RL to compensate
    for modeling inaccuracies and unplanned disturbances. 

    \subsection{Learning-based Methods for Loco-manipulation}
    \label{subsec:lblm} Significant progress has been made in training learning-based
    locomotion policies within physics-based simulators using RL~\citep{rudin2022learning, ha2024learning}. By leveraging techniques such
    as domain randomization~\citep{tobin2017domain}, policies trained in an end-to-end
    scheme have demonstrated robust performance in a range of loco-manipulation
    scenarios, including soccer dribbling~\citep{ji2023dribblebot}, object
    transferring~\citep{fu2023deep, dao2022sim} and other pedipulation tasks~\citep{arm2024pedipulate, cheng2023legs}.

    Despite these successes, RL-based approaches still face several notable limitations.
    First, the vast exploration space spanning diverse end-effector positions and object states renders the learning of meaningful manipulation behaviors sample-inefficient and reliant on complicated exploration strategies~\citep{zhang2024wococo, schwarke2023curiosity, yuan2025caiman}.
    Second, most policies are trained in conjunction with low-level joint
    proportional–derivative (PD) controllers to translate positional command into desired torques. However, this structure is often exploited by RL policies
    that output excessively large position targets to generate sufficient contact
    forces~\citep{lyu2024rl2ac}, resulting in poor controllability of both end-effector positions and 
    interaction forces~\citep{jeon2023learning, portela2024learning}, which is
    critical for precise loco-manipulation. While torque-based policies may mitigate this, it requires high-frequency updates to prevent overshooting, further
    increasing the complexity of training~\citep{chen2023learning, li2025sata}.

    To address these challenges, recent methods have explored incorporating demonstrations to guide RL
    agents and improve the motion quality~\citep{ha2024umi, sleiman2024guided, kang2023rl+, jenelten2024dtc},
    however, they still inherit the limitations by operating on joint positions. Drawing inspiration
    from prior work~\citep{xie2022glide, yang2023cajun}, our framework integrates 
    model-based WBC module to compute feedforward torque commands while complementing it with a learned policy to enhance robustness, resulting
    in an effective strategy for precise end-effector position and force control in loco-manipulation.

    
    \section{Preliminaries}
    \label{sec:preliminaries}

    In this section, we describe the preliminaries for better understanding the
    framework, including the dynamics model and foot trajectory generation strategy
    used in~\method.

    \subsection{Single Rigid Body Dynamics for Legged Robots}
    \label{subsec:srb}

    In our framework, we employ the single rigid body (SRB) dynamics to model the
    quadruped robot. It assumes the majority of the mass is concentrated in the
    base link of the robot, and all the limbs are massless and their inertial effects
    are negligible~\citep{ding2019real}. While more complex models such as centroidal
    model or whole-body model~\citep{dai2014whole, khazoom2024tailoring} can be
    in principle leveraged, they still inherit the limitations such as model inaccuracy
    and sensitivity to disturbances. We choose SRB to trade off complexity for
    computational efficiency.

\begin{figure}[t!]
        \centering
        \includegraphics[width=0.7\linewidth]{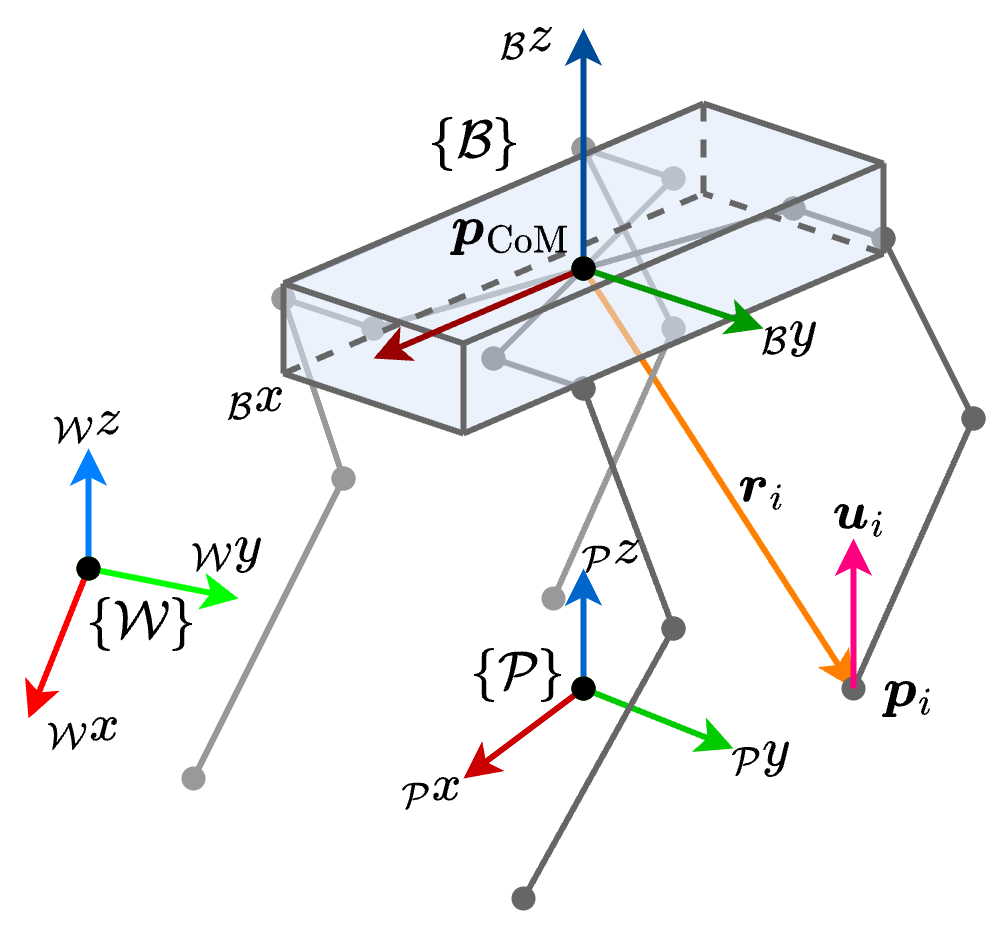}
        \caption{Illustration of the 3D single rigid body model for a quadruped
        robot. $\{\mathcal{W}\}, \{\mathcal{B}\}, \{\mathcal{P}\}$ denote the inertial world frame, the body
        frame, and the projected frame, respectively. $\bm{r}_{i}$ is the contact
        position of limb $i$ relative to the CoM. $\bm{u}_{i}$ is the external
        reaction force at the contact position.}
        \label{fig:srb}
        \vspace{-0.6cm}
    \end{figure}

    As shown in~\Cref{fig:srb}, the state of the single rigid body can be
    described by
    \begin{equation}
        \bm{x}\coloneq [\bm{p}\ \bm{v}\ \bm{R}\ {_\mathcal{B}}{\bm{\omega}}],
    \end{equation}
    where $\bm{p}\in\mathbb{R}^{3}$ is the position of the body Center of Mass (CoM);
    $\bm{v}\in\mathbb{R}^{3}$ is the CoM velocity; $\bm{R}\in SO(3)$ is the
    rotation matrix of the body frame $\{\mathcal{B}\}$ expressed in the inertial world
    frame $\{\mathcal{W}\}$; $\text{det}(\cdot)$ calculates the determinant of a matrix and
    $\mathbb{I}$ is the 3-by-3 identity matrix.
    ${_\mathcal{B}}{\bm{\omega}}\in\mathbb{R}^{3}$ is the angular velocity expressed in
    the body frame $\{\mathcal{B}\}$. Variables without subscript on the left are assumed
    to be in the inertial frame $\{\mathcal{W}\}$.
    Additionally, we define another coordinate frame, namely the \textit{projected
    frame} $\{\mathcal{P}\}$, which is centered at the projection of CoM onto the ground plane,
    and whose $x$ and $z$ axes point forward and upward, respectively.

    The input to the dynamics system is the external reaction forces $\bm{u}_{i}\in
    \mathbb{R}^{3}$ for the locations $\bm{p}_{i}\in\mathbb{R}^{3}$ in contact,
    where $i\in \{1, 2, .., N\} = \mathcal{N}$ denotes the index for
    contact locations, and $N$ is the number of contacts, as shown in~\Cref{fig:srb}.
    
    The net external wrench $\bm{\mathcal{F}}\in\mathbb{R}^{6}$ exerted on the body
    is:
    \begin{equation}
        \bm{\mathcal{F}}=
        \begin{bmatrix}
            \bm{F}    \\
            \bm{\tau}
        \end{bmatrix}
        = \sum_{i=1}^{N}
        \begin{bmatrix}
            \mathbb{I}            \\
            [\bm{r}_{i}]_{\times}
        \end{bmatrix}
        \bm{u}_{i},
    \end{equation}
    where $\bm{F}$ and $\bm{\tau}$ are the total force and torque applied at the
    CoM; $\bm{r}_{i}= \bm{p}_{i}- \bm{p}$ is the contact positions relative to CoM;
    the $[\cdot]_{\times}:\mathbb{R}^{3}\rightarrow\mathfrak{so}(3)$ operator
    converts the element from $\mathbb{R}^{3}$ to skew-symmetric matrices as
    $[\bm{a}]_{\times}\bm{b}= \bm{a}\times \bm{b}$ for all
    $\bm{a}, \bm{b}\in \mathbb{R}^{3}$. The inverse of $[\cdot]_{\times}$ is the
    vee map $[\cdot]^{\vee}:\mathfrak{so}(3) \rightarrow\mathbb{R}^{3}$ .

    The full dynamics of the rigid body can be formulated as
    \begin{equation}
        \dot{\bm{x}}=
        \begin{bmatrix}
            \dot{\bm{p}}            \\
            \dot{\bm{v}}            \\
            \dot{\bm{R}}            \\
            {_\mathcal{B}}{\dot{\bm{\omega}}}
        \end{bmatrix}
        =
        \begin{bmatrix}
            \bm{v}                                                                                              \\
            \frac{1}{M}\bm{F}+ \bm{g}                                                                           \\
            \bm{R}[{_\mathcal{B}}\bm{\omega}]_{\times}                                                                    \\
            {_\mathcal{B}}{\bm{I}}^{-1}(\bm{R}^{\top}\bm{\tau}-[{_\mathcal{B}}{\bm{\omega}}]_{\times}{_\mathcal{B}}{\bm{I}}{_\mathcal{B}}{\bm{\omega}})
        \end{bmatrix},
    \end{equation}
    where $M$ is the mass of the rigid body; $\bm{g}\in\mathbb{R}^{3}$ is the gravitational
    acceleration vector; ${_\mathcal{B}}{\bm{I}}$ is the fixed moment of inertia in the
    body frame $\{\mathcal{B}\}$.

    The dynamics of the linear and angular velocity can be further derived after
    ignoring the gyroscopic effect (assuming the angular velocity is small and its
    second-order term is negligible) and expressed in the body frame $\{\mathcal{B}\}$ as
    \begin{equation}
        \begin{bmatrix}
            {_\mathcal{B}}{\dot{\bm{v}}}      \\
            {_\mathcal{B}}{\dot{\bm{\omega}}}
        \end{bmatrix}
        =
        \begin{bmatrix}
            \frac{1}{M}\bm{R}^{\top}\bm{F}+ \bm{R}^{\top}\bm{g} \\
            {_\mathcal{B}}{\bm{I}}^{-1}\bm{R}^{\top}\bm{\tau}
        \end{bmatrix}
        = \sum_{i=1}^{N}
        \begin{bmatrix}
            \frac{1}{M}                                \\
            {_\mathcal{B}}{\bm{I}}^{-1}[{_\mathcal{B}}{\bm{r}_i}]_{\times}
        \end{bmatrix}{_\mathcal{B}}{\bm{u}_i}+ \hat{\bm{g}}, \label{eq:srb_fma}
    \end{equation}
    where $\hat{\bm{g}}= [\bm{g}, 0_{3}]^{\top}\in \mathbb{R}^{6}$ is the
    total gravity for the base.

    \subsection{Foot Trajectory Generation Strategy}
    \label{subsec:ftg} We use a heuristic strategy to generate symmetric foot trajectories for lifting and landing for each limb
    responsible for locomotion.

    Given a specific gait pattern, one can interpolate three keyframe locations $(
    {_\mathcal{P}}{\bm{p}_{\text{lift}}},{_\mathcal{P}}{\bm{p}_{\text{mid}}},{_\mathcal{P}}{\bm{p}_{\text{land}}}
    )_{i}$ expressed in the projected frame $\{\mathcal{P}\}$ for each contact leg $i$ based
    on the swing timing. The lifting location $({_\mathcal{P}}{\bm{p}_{\text{lift}}})_{i}$
    is taken from the previous contact location when the leg $i$ switches from
    contact to swing. The mid-air position $({_\mathcal{P}}{\bm{p}_{\text{mid}}})_{i}$ is located
    at the corresponding hip position $({_\mathcal{P}}{\bm{p}_{\text{hip}}})_{i}$ projected
    towards the ground in the projected frame $\{\mathcal{P}\}$ with some fixed desired foot
    height. The landing position$({_\mathcal{P}}{\bm{p}_{\text{land}}})_{i}$ is calculated
    according to the hip velocity in the projected frame $\{\mathcal{P}\}$ as
    \begin{equation}
        ({_\mathcal{P}}{\bm{p}_{\text{land}}})_{i}= ({_\mathcal{P}}{\bm{p}_{\text{hip}}})_{i}+ \frac{T_{\text{stance}}}{2}
        ({_\mathcal{P}}{\bm{v}_{\text{hip}}})_{i},
    \end{equation}
    where $T_{\text{stance}}$ is the stand duration. The desired foot trajectory
    is described using a cubic Bézier curve connecting the three keyframe positions.


    \section{Method}
    \label{sec:method} 
\begin{figure}[ht!]
        \centering
        \includegraphics[width=0.9\linewidth]{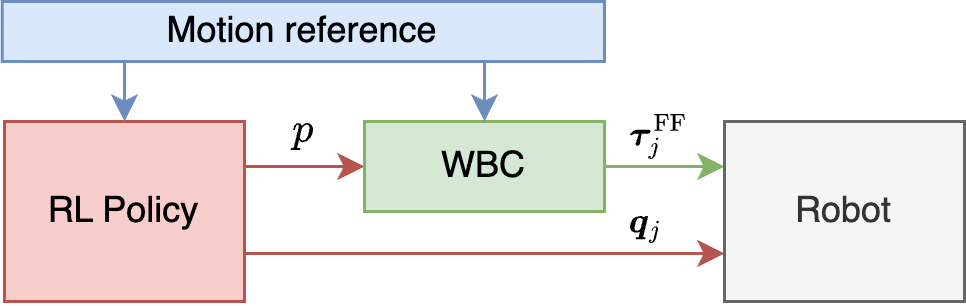}
        \caption{Overview of the \method architecture. It consists of three core components: (1) a motion reference generator, (2) a WBC module that computes feedforward joint torques $\bm{\tau}_j^{\text{FF}}$, and (3) an RL policy that generates feedback corrections to both WBC input parameters and motion reference.}
        \label{fig:method_hl}
        \vspace{-0.4cm}
    \end{figure}

As shown in~\Cref{fig:method_hl}, \method is composed of three elements:
    motion reference generation, feedforward torque acquiring through WBC, and feedback policy with RL. We describe each component in detail below.
    
    \begin{figure*}
        \centering
        \includegraphics[width=0.9\textwidth]{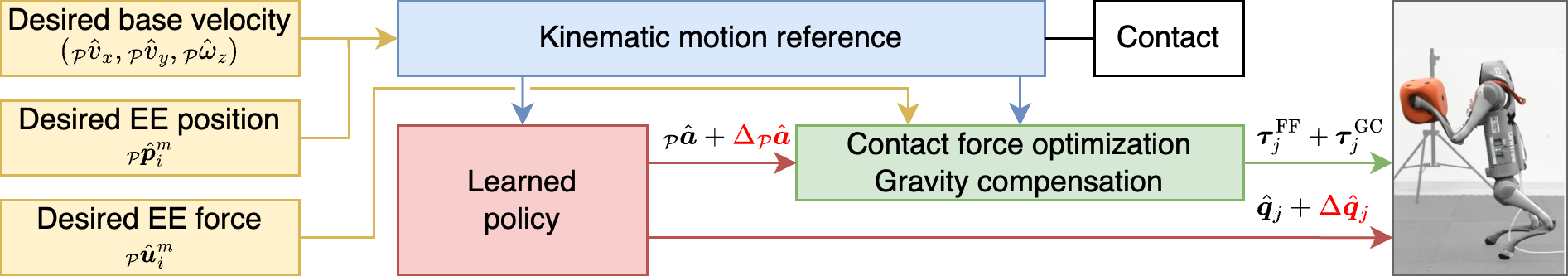}
        \caption{Detailed architecture of the \method control framework. The desired base velocity and EE positions are used to generate a kinematic motion reference, which is sent to the policy and whole-body control module. The whole-body control module also takes the desired EE force to compute the feedforward joint torques. The learned policy provides corrective feedback to the base acceleration and joint position targets, enabling robust control under modeling errors and dynamic disturbances.}
        \label{fig:method}
        \vspace{-0.6cm}
    \end{figure*}
    \subsection{Motion Reference Generation}
    \label{subsec:kinematic} For each control time step, \method starts with querying
    a motion reference for both the base and joint $({\hat{\bm{q}}},{\hat{\dot{\bm{q}}}}
    )$, where ${\hat{\bm{q}}},{\hat{\dot{\bm{q}}}}$ are the desired generalized
    coordinates and velocities respectively; ${\hat{\bm{q}}}= [{\hat{\bm{p}}}\ {\hat{\bm{R}}}
    \  \hat{\bm{q}}_{j}]$ are the desired base position, orientation and joint
    positions; ${\hat{\dot{\bm{q}}}}= [{\hat{\bm{v}}}\ \hat{\bm{\omega}}\ \hat{\dot{\bm{q}}}
    _{j}]$ are the desired base linear and angular velocities, joint velocities.
    We use the subscript $j$ to denote the joint dimensions.

    To build a framework that allows the user to interactively control the robot in a versatile manner across diverse tasks, we implement a reference generation process
    from the user command, shown in~\Cref{fig:method}. The input from the
    user includes the desired base velocity $({_\mathcal{P}}\hat{v}_{x},{_\mathcal{P}}\hat{v}_{y},{_\mathcal{P}}\hat{\omega}_{z})$, which consists of the
    desired forward, side and turning velocities, all expressed in the projected
    frame $\{\mathcal{P}\}$. Defining the reference in the projected frame $\{\mathcal{P}\}$ allows \method
    to operate with only proprioceptive information, without relying on an accurate
    estimation of coordinates in the inertial frame.

    In addition, categorizing the end-effectors $i\in\mathcal{N}$ into two kinds:
    locomotion $\mathcal{L}\subset\mathcal{N}$, marked by superscript ${l}$, and
    manipulation $\mathcal{M}\subset\mathcal{N}$, marked by superscript ${m}$, our
    kinematic reference generation also accept specification for the desired EE positions
    ${_\mathcal{P}}{\hat{\bm{p}}^m_i}\in\mathbb{R}^{3}$ for the limbs responsible for manipulation.
    By incorporating predefined gait patterns and their contact schedules, ${_\mathcal{P}}{\hat{\bm{p}}^l_i}$
    is generated by interpolating the keyframes $({_\mathcal{P}}{\bm{p}_{\text{lift}}},{_\mathcal{P}}
    {\bm{p}_{\text{mid}}},{_\mathcal{P}}{\bm{p}_{\text{land}}})_{i}$ as shown in~\Cref{subsec:ftg}.
    The reference for each joint $\hat{\bm{q}}_{j}$ is calculated using inverse
    kinematics (IK)
    \begin{equation}
        \hat{\bm{q}}_{j}= \text{IK}({_\mathcal{P}}{\bm{p}},{_\mathcal{P}}{\bm{R}}, {_\mathcal{P}}{\hat{\bm{p}}_1}, ...,{_\mathcal{P}}{\hat{\bm{p}}_N}
        ). \label{eq:ik}
    \end{equation}

    For accounting the force command and the dynamics effect from manipulation tasks, the contact
    state for manipulation
    $\hat{\bm{c}}^{m}= \{\hat{c}_{i}| i \in \mathcal{M}\}$ is always set to $1$.
    The remaining contact state of locomotion end-effector $\hat{\bm{c}}^{l}= \{\hat
    {c}_{i}| i \in \mathcal{L}\}$ is predefined according to the gait pattern.
    We fill unspecified references with either current state or zeros
    \begin{equation}
        \begin{split}
            {_\mathcal{P}}{\hat{\bm{p}}}&={_\mathcal{P}}{{\bm{p}}},{_\mathcal{P}}{\hat{\bm{R}}}={_\mathcal{P}}{{\bm{R}}}\\
            {_\mathcal{P}}{\hat{\bm{v}}}&= [{_\mathcal{P}}\hat{v}_{x},{_\mathcal{P}}\hat{v}_{y}, 0],{_\mathcal{P}}{\hat{\bm{\omega}}}
            = [0, 0,{_\mathcal{P}}\hat{\omega}_{z}], \hat{\dot{\bm{q}}}_{j}= \bm{0}.
        \end{split}
    \end{equation}

    We note that \method in principle allows any reference trajectory generation
    process including the ones using trajectory optimization or from offline
    motion library. The only requirement is to provide the contact state $\hat{c}
    _{i}\in \{0, 1\}$ for each end-effector $i \in \mathcal{N}$.
    \subsection{Generating Feedforward Torque via WBC}
    \label{subsec:ff_torque} To account for contributions from all contacts and ensure
    the robot tracks the reference motion, we employ a computationally
    lightweight whole-body controller to optimize the reaction forces leveraging the single rigid
    body model.

    We firstly calculate the base linear and angular acceleration target
    ${_\mathcal{P}}\hat{\bm{a}}= ({_\mathcal{P}}\hat{\bm{a}}_{l},{_\mathcal{P}}\hat{\bm{a}}_{a})\in\mathbb{R}^{6}$
    using a proportional-derivative controller
    \begin{align}
        {_\mathcal{P}}\hat{\bm{a}}_{l} & = \kappa_{lp}({_\mathcal{P}}{\hat{\bm{p}}}-{_\mathcal{P}}{{\bm{p}}}) + \kappa_{ld}({_\mathcal{P}}{\hat{\bm{v}}}-{_\mathcal{P}}{{\bm{v}}})                                       \\
        {_\mathcal{P}}\hat{\bm{a}}_{a} & = \kappa_{ap}\text{log}({_\mathcal{P}}{\hat{\bm{R}}}^{\top}\cdot{_\mathcal{P}}{{\bm{R}}})^{\vee}+ \kappa_{ad}({_\mathcal{P}}{\hat{\bm{\omega}}}-{_\mathcal{P}}{{\bm{\omega}}}),
    \end{align}
    where $\kappa_{lp}, \kappa_{ld}, \kappa_{ap}, \kappa_{ad}$ are the linear
    and angular, proportional and derivative gains respectively;
    $\text{log}(\cdot)^{\vee}:SO(3) \rightarrow\mathbb{R}^{3}$ converts a
    rotation matrix into an angle-axis representation, which is a vector in $\mathbb{R}
    ^{3}$.

    In addition to the desired acceleration ${_\mathcal{P}}\hat{\bm{a}}$, \method also accept
    user's specification of desired EE forces ${_\mathcal{P}}\hat{\bm{u}}_{i}$ for the
    manipulation end-effectors $i\in \mathcal{M}$, as shown in~\Cref{fig:method}. Using ${_\mathcal{B}}\hat{\bm{a}}$ and ${_\mathcal{B}}\hat{\bm{u}}_{i}$ in the base frame
    $\{\mathcal{B}\}$, we
    formulate the following QP to optimize the reaction force
    \begin{subequations}
        \begin{equation}
            \min_{{_\mathcal{B}}\bm{u}_i, i\in\mathcal{N}}\norm{\Delta{_\mathcal{B}}{\bm{a}}}_{\bm{U}}
            ^{2}+ \sum_{i\in \mathcal{M}}\norm{\Delta{_\mathcal{B}}\bm{u}_i}_{\bm{V}}^{2}+
            \sum_{i\in \mathcal{L}}\norm{{_\mathcal{B}}\bm{u}_i}_{\bm{W}}^{2}
        \end{equation}
        \vspace{-0.3cm}
        \begin{alignat}
            {2}\text{subject}\ \text{to}:{_\mathcal{B}}{\bm{a}} & = \sum_{i=1}^{N}\bm{A}_{i}{_\mathcal{B}}\bm{u}_{i}+ \hat{\bm{g}} \\
            ({_\mathcal{B}}\bm{u}_{i})_{z}                      & = 0,                                                  &  & i\in \mathcal{L}_{\text{swing}}  \\
            \norm{({_\mathcal{B}}\bm{u}_{i})_{x}}               & \leq \mu({_\mathcal{B}}\bm{u}_{i})_{z},                         &  & i\in \mathcal{L}_{\text{stance}} \\
            \norm{({_\mathcal{B}}\bm{u}_{i})_{y}}               & \leq \mu({_\mathcal{B}}\bm{u}_{i})_{z},                         &  & i\in \mathcal{L}_{\text{stance}} \\
            u_{z, \min}                               & \leq({_\mathcal{B}}\bm{u}_{i})_{z}\leq u_{z, \max}, \quad       &  & i\in \mathcal{L}_{\text{stance}}
        \end{alignat}
    \end{subequations}
    where $\Delta{_\mathcal{B}}{\bm{a}}={_\mathcal{B}}\hat{\bm{a}}-{_\mathcal{B}}{\bm{a}}$, $\Delta{_\mathcal{B}}\bm{u}_{i}
    ={_\mathcal{B}}\hat{\bm{u}}_{i}-{_\mathcal{B}}\bm{u}_{i}$;
    ${_\mathcal{B}}{\bm{a}}= [{_\mathcal{B}}{\dot{\bm{v}}},{_\mathcal{B}}{\dot{\bm{\omega}}}]\in\mathbb{R}^{6}$
    is the acceleration of the single rigid body; $\bm{A}_{i}$ is the generalized
    inverse inertia matrix defined in~\Cref{eq:srb_fma}; $\mu$ is the friction
    coefficient; $\mathcal{L}_{\text{swing}}, \mathcal{L}_{\text{stance}}$ are the
    set of end-effectors for locomotion in swing and stance respectively;
    $\bm{U}, \bm{V}, \bm{W}\succ 0$ are positive definite weight matrices. We note
    that friction checking is only performed on locomotion end-effectors due to the
    uncertainty of contact surface for manipulation. Leveraging the SRB model, \method
    efficiently accounts for the dynamic effects from contacts.

    The feedforward joint torques $\bm{\tau}^{\text{FF}}_{j}$ are calculated using
    \begin{equation}
        \bm{\tau}^{\text{FF}}_{j}= \sum_{i = 1}^{N}\bm{J}_{i}^{\top}\cdot{_\mathcal{B}}\bm{u}
        _{i},
    \end{equation}
    where $\bm{J}_{i}$ is the Jacobian corresponds to the end-effector $i$.

    In addition to the feedforward torque from the reaction force optimization module,
    we calculate an additional torque term $\bm{\tau}^{\text{GC}}_{j}$ to
    compensate the gravity and account for the limb inertia, for each joint $k$
    \begin{equation}
        (\bm{\tau}^{\text{GC}}_{j})_{k}= -\sum_{l \in \mathcal{D}(k)}\bm{J}_{lk}^{\top}
        \cdot m_{l}\bm{g},
    \end{equation}
    where $\bm{J}_{lk}$ is the Jacobian matrix mapped from the CoM velocity of link
    $l$ to joint $k$; $m_{l}$ is the mass of link $l$, and $\mathcal{D}(k)$ is a
    set of descendant links of $k$.

    \subsection{Learned Policy}
    \label{subsec:learned_pi} Directly applying the feedforward torque
    $\bm{\tau}_{j}$ may not be enough to accomplish complex loco-manipulation
    tasks due to the large model mismatch. As a remedy, \method incorporate a
    learned policy trained in simulated environments using
    RL to improve the overall robustness of the controller
    over unconsidered dynamic effects.

    An RL problem is formulated as a Markov Decision Process (MDP), represented by
    a tuple $\langle\mathcal{O}, \mathcal{A}, P, r, \gamma\rangle$, where
    $\mathcal{O}$ is the observation space; $\mathcal{A}$ is the action space;
    $P_{O'|O, A}$ is the transition probability; $r:\mathcal{O}\times\mathcal{A}\rightarrow
    \mathbb{R}$ is the reward function; $\gamma$ is the discounted factor. The
    policy $\pi_{\theta}:\mathcal{O}\rightarrow\mathcal{A}$, which is
    parameterized by $\theta$, is trained to maximize the expected sum of discounted
    reward
    \begin{equation}
        \mathbb{E}_{o_0\sim p_0(\cdot), o_{t+1}\sim P(\cdot|o_t, \pi_{\theta}(o_t))}
        \sum_{t=0}^{T}\gamma^{t}r(o_{t}, \pi_{\theta}(o_{t})),
    \end{equation}
    over an episode of $T$, where $p_{0}$ is the initial state distribution.

    We design the observation space $\mathcal{O}$ to include the proprioceptive
    information, gait information, kinematic joint position target, and user commands.
    They are chosen to ensure reward function can be successfully induced from
    only the observation. We stack 6-step history of observations as input to the policy~\citep{margolis2023walk}.
    In contrast to previous works~\citep{xie2022glide, yang2023cajun},
    the action $a_{t}\in\mathcal{A}$ is designed to have two separate heads:
    base acceleration correction $\Delta{_\mathcal{P}}{\hat{\bm{a}}}$ and joint position
    correction $\Delta \hat{\bm{q}}_{j}$, providing feedback to both feedforward torque
    calculation and joint positions. The surrogate targets are
    \begin{equation}
        \begin{aligned}
            {_\mathcal{P}}{\Tilde{\bm{a}}} & ={_\mathcal{P}}{\hat{\bm{a}}}+\Delta{_\mathcal{P}}{\hat{\bm{a}}}  \\
            \Tilde{\bm{q}}_{j}   & = \hat{\bm{q}}_{j}+ \Delta\hat{{\bm{q}}}_{j},
        \end{aligned}
    \end{equation}
    where ${_\mathcal{P}}{\Tilde{\bm{a}}}$ is taken as the target for the base
    acceleration for the reaction force optimization module. The desired joint position
    $\Tilde{\bm{q}}_{j}$ is used to calculate the final joint torque command sent
    to the robot, formulated as
    \begin{equation}
        \bm{\tau}_{j}= \bm{\tau}^{\text{FF}}_{j}+\bm{\tau}^{\text{GC}}_{j}+ k_{p}
        (\Tilde{\bm{q}}_{j}- \bm{q}_{j})+ k_{d}(\dot{\hat{\bm{q}}}_{j}- \dot{\bm{q}}
        _{j}),
    \end{equation}
    where $k_{p}, k_{d}$ are the proportional and derivative gains for the joint
    PD controller.

    The reward function consists of a combination of task-related rewards and
    regularizations
    \begin{equation}
        r = r_{\text{task}}+ r_{\text{reg}},
    \end{equation}
    where $r_{\text{task}}= \prod_{i}r^{i}_{\text{task}}$ and
    $r_{\text{reg}}= \prod_{j}r^{j}_{\text{reg}}$ are a product of series sub-rewards.
    Both rewards are designed to ensure the success of tracking user commands and
    regularized action. For detailed description of the observation space and reward
    functions, please refer to~\Cref{tab:state} and~\Cref{tab:reward}.

    \begin{table}[ht!]
        \centering
        \begin{tabular}{lll}
            \thickhline Term name  & Symbol                                                   & Dimension              \\
            \hline
            Base height            & $\bm{p}_{z}$                                             & 1                      \\
            Projected gravity      & ${_\mathcal{B}}\bm{g}$                                             & 3                      \\
            Base linear velocity   & ${_\mathcal{B}}\bm{v}$                                             & 3                      \\
            Base angular velocity  & ${_\mathcal{B}}\bm{\omega}$                                        & 3                      \\
            Joint position         & $\bm{q}_{j}- \bm{q}_{j0}$                                & 12                     \\
            Joint velocity         & $\dot{\bm{q}}_{j}$                                       & 12                     \\
            Gait phase             & $\bm{\phi}$                                              & 4                      \\
            Gait mode              & $\bm{\psi}$                                              & 4                      \\
            Desired joint position & $\hat{\bm{q}}_{j}- \bm{q}_{j0}$                          & 12                     \\
            Velocity command       & $({_\mathcal{P}}\hat{v}_{x},{_\mathcal{P}}\hat{v}_{y},{_\mathcal{P}}\hat{\omega}_{z})$ & 3                      \\
            EE position command    & ${_\mathcal{P}}{\hat{\bm{r}}^m_i}$                                 & $3 \cdot|\mathcal{M}|$ \\
            EE force command       & ${_\mathcal{P}}\hat{\bm{u}}_{i}$                                   & $3 \cdot|\mathcal{M}|$ \\
            Last action            & $a_{t-1}$                                                & 18                     \\
            \thickhline
        \end{tabular}
        \caption{Detailed observation space $\mathcal{O}$. $\bm{q}_{j0}$ is the default
        joint position; $\bm{\phi}\in[-1, 1]$ is the phase for each periodic
        limb motion; $\bm{\psi}\in\{-1, 0, 1\}$ is the mode for each limb to distinguish
        the periodic patterns (\textit{swing}, \textit{gait}, \textit{stance}); $|
        \cdot|$ is the cardinality of the set.}
        \label{tab:state}
        \vspace{-0.6cm}
    \end{table}

    \begin{table}[ht!]
        \centering
        \begin{tabular}{l|lll}
            \thickhline Reward                 & Term name          & Error $\bm{\epsilon}$                             & Sensitivity $\sigma$ \\
            \hline
            \multirow{5}{*}{$r_{\text{task}}$} & Base height        & $\bm{p}_{z}- \hat{\bm{p}}_{z}$                    & 0.1 / 0.2            \\
                                               & Base orientation   & ${_\mathcal{B}}\bm{g}_{z}-{_\mathcal{B}}\hat{\bm{g}}_{z}$             & 0.3 / 0.6            \\
                                               & Linear velocity    & ${_\mathcal{P}}{\bm{v}}_{x,y}-{_\mathcal{P}}\hat{\bm{v}}_{x,y}$       & 0.2 / 0.3            \\
                                               & Angular velocity   & ${_\mathcal{P}}{\bm{\omega}}_{z}-{_\mathcal{P}}\hat{\bm{\omega}}_{z}$ & 0.3 / 0.4            \\
                                               & EE position        & ${_\mathcal{P}}{{\bm{r}}^m_i}-{_\mathcal{P}}{\hat{\bm{r}}^m_i}$       & 0.1                  \\
            \hline
            \multirow{5}{*}{$r_{\text{reg}}$}  & Contact mismatch   & $\hat{c}_{i}\neq c_{i}$                           & -                    \\
                                               & Joint acceleration & $\ddot{\bm{q}}_{j}$                               & 700 / 500            \\
                                               & Joint torque       & $\bm{\tau}_{j}$                                   & 100                  \\
                                               & Action rate        & $a_{t}- a_{t-1}$                                  & 10.0                 \\
                                               & Action scale       & $a_{t}$                                           & 8.0                  \\
            \thickhline
        \end{tabular}
        \caption{Detailed reward functions. To ensure all reward functions produce
        values with the range of $[0, 1]$, we map the error term for each reward
        using $r = \exp(\norm{\bm{\epsilon}}^{2}/ \sigma^{2})$. The contact mismatch
        reward is mapped using $r = 0.5^{|\hat{c}_i \neq c_i|}$.
        $(\cdot)/(\cdot)$ indicates different values for quadruped or biped
        tasks. We set desired base height $\hat{\bm{p}}_{z}$ to 0.3 and 0.45,
        desired gravity vector ${_\mathcal{B}}\hat{\bm{g}}_{z}$ to $[0, 0, -1]$ and
        $[-1, 0, 0]$ for quadruped and biped tasks.}
        \label{tab:reward}
        \vspace{-0.6cm}
    \end{table}
    \section{Results}
    \label{sec:results} \method offers a general framework for whole-body loco-manipulation
    on legged systems. We demonstrate its effectiveness on the Unitree Go2~\citep{go2},
    a small-scale quadruped robot, across a variety of tasks involving both
    quadrupedal and bipedal locomotion.

    We implemented two scenarios targeting the quadrupedal and bipedal tasks, respectively.
    In the quadruped tasks, the robot walks using three legs while lifting the front-left
    leg to perform manipulation. For the bipedal tasks, it walks solely on its hind
    legs while using both front legs for manipulation. We design the base
    orientation to keep flat to the ground for quadruped tasks, while
    demonstrating more challenging loco-manipulation tasks by making the robot to
    perform bipedal walking with upright pose. These bipedal demonstrations
    highlight the potential of \method for applications on humanoids.

    We leverage Isaac Lab~\citep{mittal2023orbit}, a massive parallel training
    framework on GPU, to efficiently train the policy with Proximal
    Policy Optimization~\citep{schulman2017proximal}. The detailed training
    hyperparameters can be found in~\Cref{tab:train}. During training, we
    leverage qpth~\citep{amos2017optnet}, a fast batch QP solver implemented in
    PyTorch, to solve parallel QPs to generate feedforward torques. Despite
    the effectiveness of qpth in training, we employ OSQP~\citep{stellato2020osqp}
    as a faster QP solver for a single problem to ensure the whole control
    pipeline runs at 100 Hz in the real-world experiments.

    To facilitate the training with force command at end-effectors, we apply virtual external forces acted at the same end-effector in the opposite direction, similarly
    to the training technique proposed by~\citet{portela2024learning}. We employ various
    Domain Randomization~\citep{tobin2017domain} to ensure successful sim-to-real
    transfer. Detailed command sampling and domain randomization can be found in~\Cref{tab:dr}.
    \begin{table}[ht!]
        \centering
        \begin{tabular}{p{0.25\linewidth}l|ll}
            \thickhline Term                                                & Value                 & Term                 & Value           \\
            \hline
            \# environments                                                 & 4096                  & \# steps per iter    & 24              \\
            Episode length                                                  & 10 (s)                & $\gamma$             & 0.99            \\
            Learning rate                                                   & $1e^{-3}$             & $\lambda$            & 0.95            \\
            Desired KL                                                      & 0.02                  & Clip ratio           & 0.2             \\
            Value loss coeff                                                & 1.0                   & Policy network       & MLP             \\
            Entropy coeff                                                   & $1e^{-3}$             & Policy hidden        & [512, 256, 128] \\
            \multirow{2}{\linewidth}{Action head scale (base acceleration)} & \multirow{2}{*}{5.0}  & Policy activation    & ELU             \\
                                                                            &                       & Value network        & MLP             \\
            \multirow{2}{\linewidth}{Action head scale (joint position)}    & \multirow{2}{*}{0.15} & Value hidden         & [512, 256, 128] \\
                                                                            &                       & Value activation     & ELU             \\
            Joint P gain $k_{p}$                                            & 40                    & Joint D gain $k_{d}$ & 1.0             \\
            \thickhline
        \end{tabular}
        \caption{Detailed training hyperparameters.}
        \label{tab:train}
        \vspace{-0.6cm}
    \end{table}

    \begin{table}[ht!]
        \vspace{0.2cm}
        \centering
        \begin{tabular}{ll}
            \thickhline Term                                 & Value                                \\
            \hline
            Forward velocity ${_\mathcal{P}}\hat{v}_{x}$               & $[-0.5, 0.5]$ (m/s)                  \\
            Side velocity ${_\mathcal{P}}\hat{v}_{y}$                  & $[-0.5, 0.5]$ / $[-0.0, 0.0]$ (m/s)  \\
            Turning velocity ${_\mathcal{P}}\hat{\omega}_{z}$          & $[-0.5, 0.5]$ (rad/s)                \\
            EE position $x$ $({_\mathcal{P}}{\hat{\bm{p}}^m_i})_{x}$   & $[0.1934, 0.5]$ / $[0.15, 0.30]$ (m) \\
            EE position $y$ $({_\mathcal{P}}{\hat{\bm{p}}^m})_{y}$     & $[0, 0.2]$ (m)                       \\
            EE position $z$ $({_\mathcal{P}}{\hat{\bm{p}}^m})_{z}$     & $[0.0, 0.4]$ / $[0.3, 0.9]$ (m)      \\
            EE force in $x, y, z$ ${_\mathcal{P}}\hat{\bm{u}}$         & $[-30, 30]$ / $[-20, 20]$ (N)        \\
            Friction coefficient                             & $[0.5, 1.5]$                         \\
            Add mass for base                                & $[-2.0, 2.0]$ (kg)                   \\
            CoM offset for base                              & $[-0.05, 0.05]$ (m)                  \\
            Actuator random delay                            & $[0, 20]$ (ms)                       \\
            Base position noise $\bm{p}_{x}, \bm{p}_{y}$     & $[-0.05, 0.05]$ (m)                  \\
            Base position noise $\bm{p}_{z}$                 & $[-0.02, 0.02]$ (m)                  \\
            Base orientation noise $\bm{R}$                  & $[-0.05, 0.05]$                      \\
            Base linear velocity noise $\bm{v}$              & $[-0.1, 0.1]$ (m/s)                  \\
            Base angular velocity noise $\bm{\omega}$        & $[-0.15, 0.15]$ (m/s)                \\
            Joint position noise $\bm{q}_{j}$                & $[-0.01, 0.01]$ (rad)                \\
            Joint velocity noise $\dot{\bm{q}}_{j}$          & $[-1.5, 1.5]$ (rad/s)                \\
            \thickhline
        \end{tabular}
        \caption{Detailed command sampling and domain randomization. $(\cdot) / (
        \cdot)$ represents different values used for quadruped and biped tasks respectively.
        The orientation noise is estimated from the noise sampled on unit
        quaternion representations. Besides the terms listed above, we also add
        random push on the robot base within an episode. Note that the EE position
        and forces are listed only for the front-left limb of the robot. For
        biped tasks, we sample the quantities in symmetry for the front-right limb.}
        \label{tab:dr}
        \vspace{-0.6cm}
    \end{table}

    \subsection{Quantitative Evaluation}
    \label{subsec:quantitative} To evaluate the performance of \method, we compare
    \method with the following baselines in simulated environments in terms of tracking the desired base velocity, desired EE positions and forces:
    \begin{itemize}
        \item \textbf{Vanilla}: Policies trained to track the commands in an end-to-end fashion~\citep{portela2024learning}. We use the same contact information from the gait pattern to facilitate the policy to generate proper gait patterns;

        \item \textbf{Imitation}: Policies trained to track the target joint angles from kinematic reference~\citep{peng2018deepmimic} in addition to the vanilla policies;
        \item \textbf{Residual}: Policies trained to produce joint position residuals in addition to the kinematic reference, without WBC to generate feedforward torques;

        \item \textbf{\method-base}: WBC with a feedback policy outputting acceleration correction
            $\Delta{_\mathcal{P}}{\hat{\bm{a}}}$ only;

        \item \textbf{\method-joint}: WBC and a feedback policy outputting joint correction
            $\Delta\hat{{\bm{q}}}_{j}$ only;

        \item \textbf{\method-ff}: WBC only (no training needed).
    \end{itemize}
    Note that the action space of the vanilla and imitation policies
            is an offset of joint positions relative to the fixed default positions.
    We set the action scale to 0.25 to facilitate exploration, a common choice in the previous works~\citep{rudin2022learning}. In comparison, \method and residual policies have joint actions relative to the kinematic reference with a scale of 0.15. 
    
    During evaluation, we randomly sample user commands in base velocity, EE positions and forces. 
    As shown in~\Cref{tab:quantitative}, \method achieves comparable or superior
    performance to all baselines across both quadrupedal and bipedal tasks. Notably,
    our method exhibits a clear advantage in tracking target end-effector
    positions, significantly reducing tracking errors. These results highlight \method’s precision and effectiveness in whole-body loco-manipulation
    for both locomotion modes. Note that since we apply virtual external forces at the end-effectors, the baselines’ lower performance in tracking end-effector positions indicates that they fail to generate the appropriate desired forces while following user-commanded end-effector positions.

    We trained vanilla and imitation policies using the similar reward structure to those of \method. While we performed reward shaping as best as we could, we found it difficult to balance the various objectives. For those achieving good tracking behavior in velocity and EE position and forces, they often produced much less regularized actions for deployment.
    We also emphasize the critical role of corrective feedback through the learned
    policy in \method. As shown in~\Cref{tab:quantitative}, incorporating residual feedback, particularly at the joint position level, leads to a substantial reduction in EE tracking error in comparison with \method-ff. 
    While the residual policies were able to track certain commands in quadruped tasks without WBC, they failed to generate sufficient contact forces for bipedal walking, further highlighting the importance of feedforward torques.

    \begin{table*}
        [t!]
        \centering
        \begin{tabular}{l|p{2.2cm}p{2.2cm}p{2.2cm}|p{2.2cm}p{2.2cm}p{2.2cm}}
            \thickhline      & \multicolumn{3}{c|}{Error in tracking from quadruped task}             & \multicolumn{3}{c}{Error in tracking from biped task}                   \\
                             & lin vel (\text{m/s}) $\downarrow$ &  ang vel (\text{rad/s}) $\downarrow$ &  EE pos (\text{m}) $\downarrow$ &  lin vel (\text{m/s}) $\downarrow$ &  ang vel (\text{rad/s}) $\downarrow$ &  EE pos (\text{m}) $\downarrow$ \\
            \hline
            Vanilla              & $0.387 \pm 0.022$                              & $0.257 \pm 0.026$                               & $0.254 \pm 0.009$                          & $0.313 \pm 0.026$                             & $0.353 \pm0.066$                               & $0.431 \pm 0.024$                         \\
            Imitation        & $0.383 \pm 0.022$                              & $0.253 \pm 0.025$                               & $0.257 \pm 0.010$                          & $0.310 \pm 0.027$                             & $0.334 \pm 0.060$                              & $0.448 \pm 0.018$                         \\
            Residual              & $0.153 \pm 0.027$                              & $0.153 \pm 0.067$                               & $0.077 \pm 0.013$                          & $0.383 \pm 0.054$                             & $0.508 \pm0.271$                               & $0.347 \pm 0.030$                         \\
            \method-base     & $0.321 \pm 0.041$                              & $0.293 \pm 0.126$                               & $0.168 \pm 0.036$                          & $0.305 \pm 0.060$                             & $0.389 \pm 0.171$                              & $0.453 \pm 0.025$                         \\
            \method-joint    & $0.108 \pm 0.014$                              & $0.101 \pm0.032$                                & $0.046 \pm 0.007$                          & $0.306 \pm 0.046$                             & $0.383 \pm 0.109$                              & $0.134 \pm 0.044$                         \\
            \method-ff       & $0.374 \pm 0.036$                              & $0.404 \pm 0.154$                               & $0.286 \pm 0.048$                          & $0.320 \pm 0.061$                             & $0.389 \pm 0.187$                              & $0.447 \pm 0.026$                         \\
            \textbf{\method} & $\mathbf{0.087 \pm 0.009}$                     & $\mathbf{0.085 \pm 0.022}$                      & $\mathbf{0.039 \pm 0.003}$                 & $0.286 \pm 0.039$                             & $0.352 \pm 0.075$                              & $\mathbf{0.036 \pm 0.002}$                \\
            \thickhline
        \end{tabular}
        \caption{Quantitative evaluation of \method compared with baselines. The
        mean and variance are calculated across 3 different seeds with 1000 episode
        for each seed. For biped tasks, the end-effector tracking error is calculated
        as the mean of tracking FL and FR end-effectors.}
        \label{tab:quantitative}
        \vspace{-0.6cm}
    \end{table*}

    \subsection{Real-world Experiments}
    \label{subsec:realworld}
    \begin{figure}[b!]
        \vspace{-0.2cm}
        \centering
        \begin{subfigure}
            {0.48\linewidth}
            \centering
            \includegraphics[width=\linewidth,]{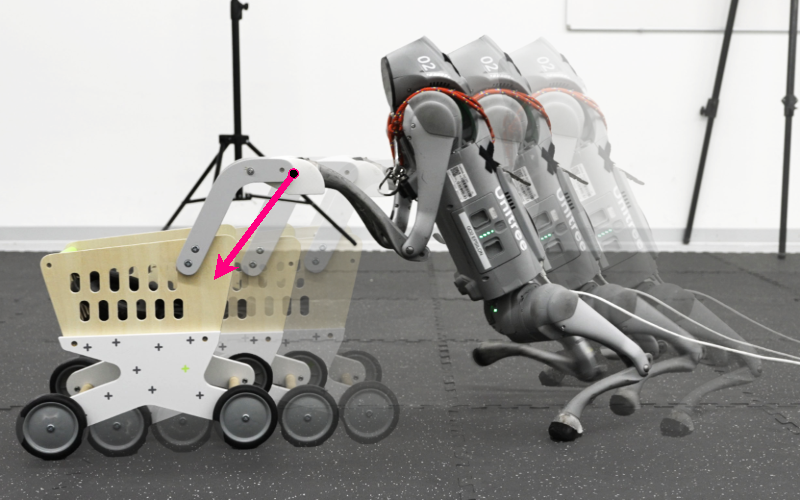}
        \end{subfigure}
        \begin{subfigure}
            {0.48\linewidth}
            \centering
            \includegraphics[width=\linewidth,]{
                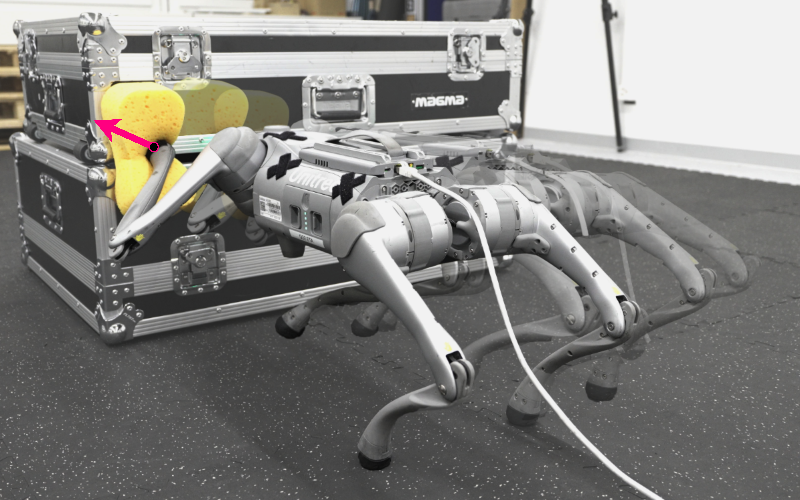
            }
        \end{subfigure}
        \caption{Snapshots of two whole-body loco-manipulation tasks, where the desired
        EE force are overlaid as pink arrows. \textit{Upper}: pushing a shopping
        cart while walking in bipedal mode; \textit{bottom}: holding a sponge while
        walking in quadrupedal mode.}
        \label{fig:hw}
        \vspace{-0.2cm}
    \end{figure}

    \begin{figure}[b!]
        \centering
        \begin{subfigure}
            {0.32\linewidth}
            \centering
            \includegraphics[width=\linewidth,]{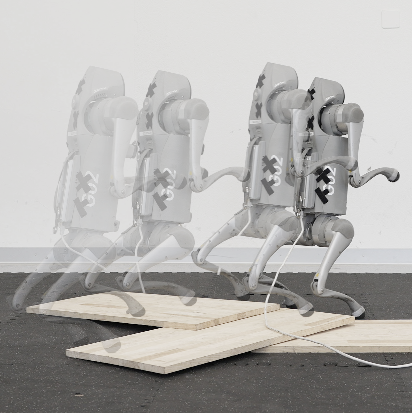}
        \end{subfigure}
        \begin{subfigure}
            {0.64\linewidth}
            \centering
            \includegraphics[width=\linewidth]{
                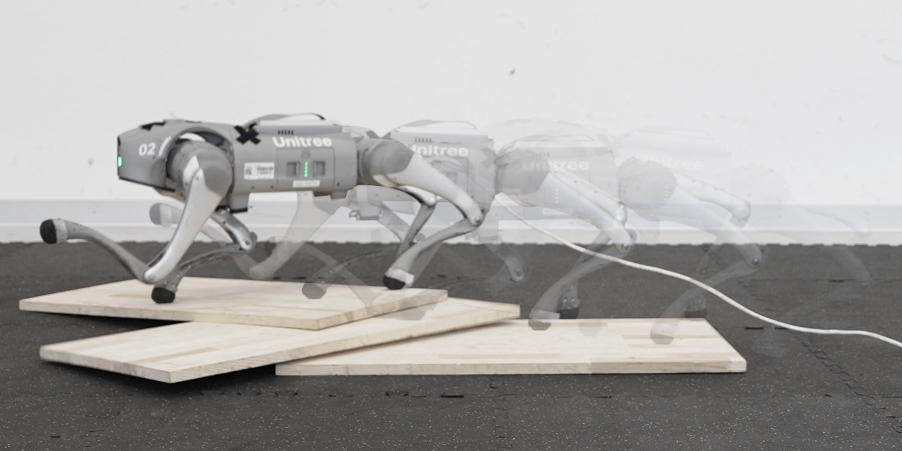
            }
        \end{subfigure}
        \caption{Snapshots of experiments demonstrating \method's robustness on uneven terrains in bipedal mode (\textit{left})
        and quadrupedal mode (\textit{right}).}
        \label{fig:robust}
        \vspace{-0.2cm}
    \end{figure}

    By changing user inputs during runtime, we demonstrate the success execution of diverse loco-manipulation skills such as bipedal cart pushing and dice holding with the same policy trained for bipedal tasks. Similarly with the policy trained for quadrupedal tasks, \method achieves stable object holding and plate balancing while walking with other three legs, as shown in the snapshots of these experiments are included in~\Cref{fig:teaser}. In more
    detail, we overlay the multiple images of shopping cart pushing and sponge
    holding tasks in~\Cref{fig:hw} to demonstrate that \method enables the quadruped
    to apply the desired force stably while walking dynamically. 

    In addition, we demonstrate the robustness of \method by commanding the robot on uneven terrains and exerting external pushes in both bipedal and quadrupedal tasks, shown in~\Cref{fig:robust}. 
    We refer the
    interested readers to the hardware demonstrations in the supplementary video.

    \subsection{Further Comparison with Vanilla Policy}
    \label{subsec:comparison}
    \begin{figure}[t!]
        \vspace{0.2cm}
        \centering
        \begin{subfigure}
            {.48\linewidth}
            \centering
            \includegraphics[width=\linewidth]{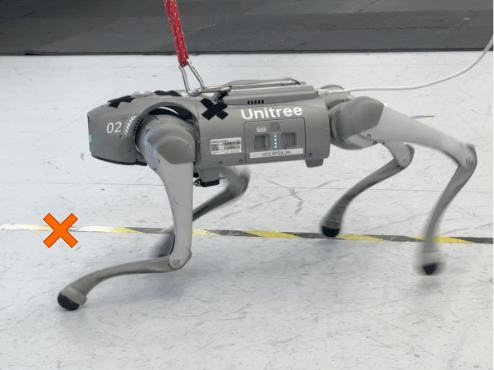}
        \end{subfigure}%
        \vspace{0.05cm}
        \begin{subfigure}
            {.48\linewidth}
            \centering
            \includegraphics[width=\linewidth]{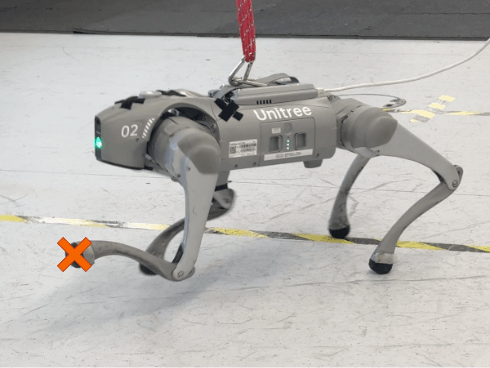}
        \end{subfigure}
        \caption{Comparison between vanilla policy (\textit{left}) and \method
        (\textit{right}).}
        \label{fig:e2evsdaffi}
        \vspace{-0.4cm}
    \end{figure}
    We further compare the performance of \method against a vanilla policy which is regularized and deployable on hardware in terms of tracking the desired EE position. 
    While both policies achieve smooth end-effector position control during static standing, 
    \method exhibits more accurate EE tracking while walking with the remaining three legs, as shown in~\Cref{fig:e2evsdaffi}. In contrast, the vanilla policy tends to
    sacrifice EE tracking precision in favor of maintaining overall stability
    during dynamic locomotion.

    We attribute this discrepancy to the relatively large exploration space in the vanilla training setup, which makes it difficult for the policy to consistently prioritize accurate EE tracking while generating appropriate contact forces through its end-effectors.
    Instead, \method separates feedforward torque from WBC and joint feedback from policy, which allows the learned policy to provide corrective offsets relative to reference positions with a much smaller exploration space, improving robustness without significantly altering the desired motion.

    \subsection{Compliance}
    \label{subsec:compliance}

\begin{figure}[t!]
        \vspace{0.2cm}
        \centering
        \begin{subfigure}
            {.48\linewidth}
            \centering
            \includegraphics[width=\linewidth]{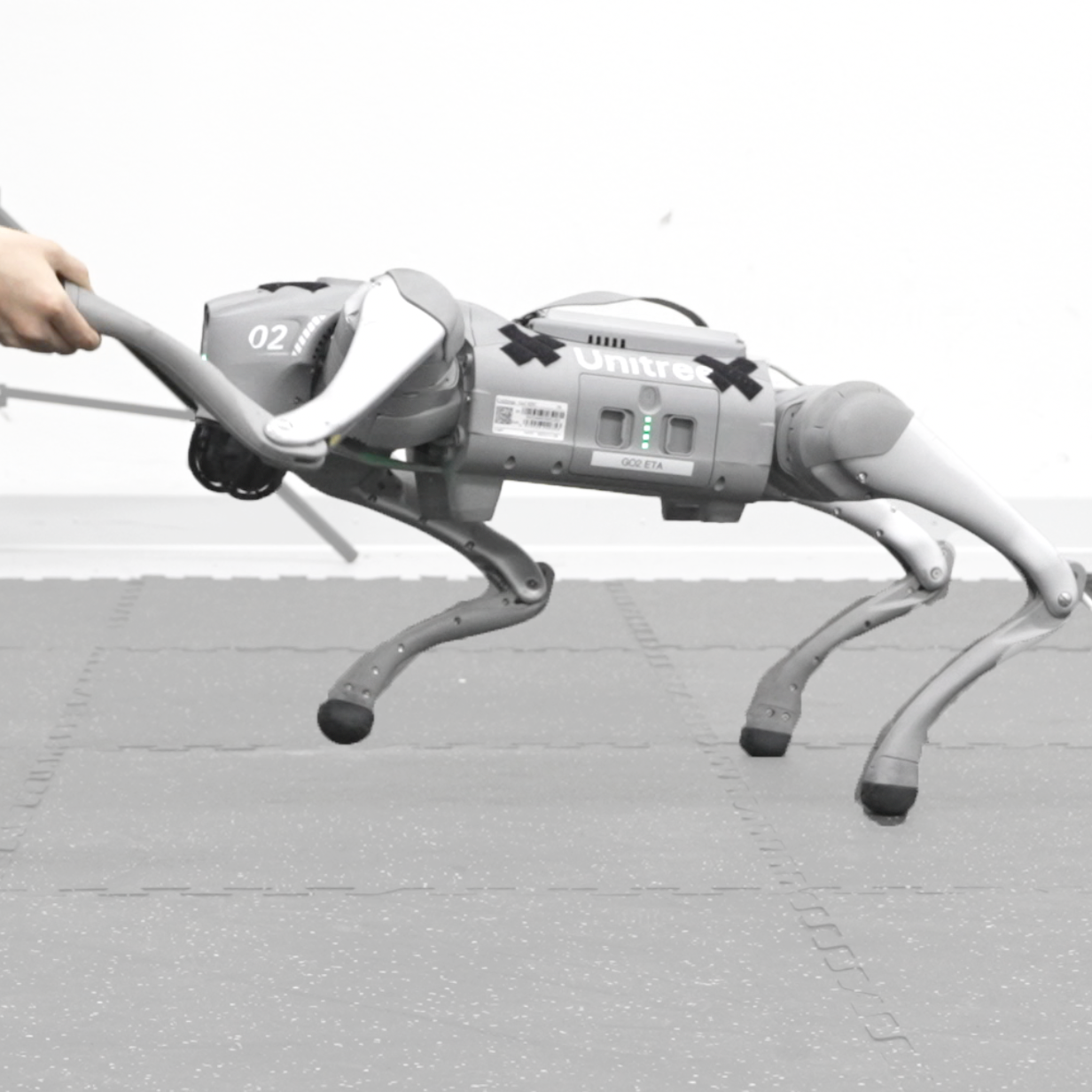}
        \end{subfigure}%
        \vspace{0.05cm}
        \begin{subfigure}
            {.48\linewidth}
            \centering
            \includegraphics[width=\linewidth]{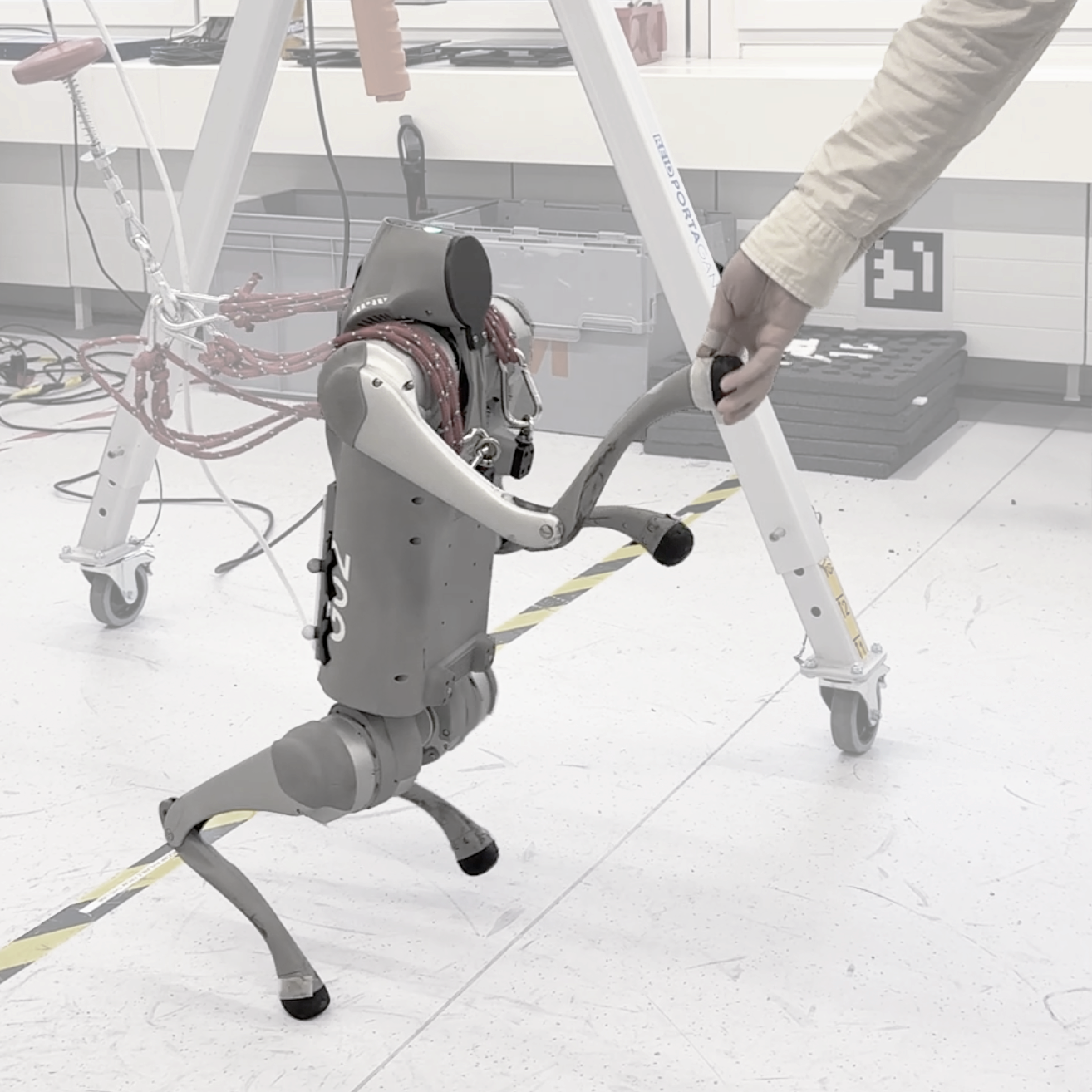}
        \end{subfigure}
        \caption{Compliance enabled by \method in quadruped mode (\textit{left})
        and bipedal mode (\textit{right}). The robot is commanded to maintain its
        end-effector position while allowing external forces to displace it compliantly.}
        \label{fig:compliance}
        \vspace{-0.6cm}
    \end{figure}

    Finally, we showcase one of the features enabled by \method, compliance, by
    lowering the PD gains at the joints associated with the manipulation end-effectors.
    Thanks to the WBC, the robot is able to maintain a stable
    end-effector position while walking and being compliant against external pushes.
    As illustrated in~\Cref{fig:compliance}, this compliance is demonstrated through
    an interactive handshake, where users are able to physically engage with the
    robot safely. By decoupling the feedforward torque and PD feedback, \method
    enables a flexible trade-off between compliance and accuracy in end-effector
    tracking, an essential property for ensuring safe and adaptive interactions.

    \section{Conclusion}
    We present \method, a hybrid control framework that combines a model-based whole-body controller with a learned policy to enable robust
    and precise whole-body loco-manipulation on legged robots. By leveraging a computationally efficient QP based on the SRB model, \method optimizes feedforward torque commands while maintaining robustness through learning-based feedback. Our results in both simulation and on hardware
    demonstrate \method's advantage in tracking user commands across a range of quadrupedal and bipedal loco-manipulation tasks. Additionally, the framework allows for a
    flexible trade-off between tracking accuracy and compliance, which is crucial for safe
    and adaptive interaction with environment. 

    \method is not without limitations. Currently, it relies solely on proprioceptive information, which negatively affects performance due to drift in state estimation. As a next step, we aim to incorporate additional sensing modalities to enhance robustness and accuracy.
    Nevertheless, we see strong potential for \method in future loco-manipulation research, including the integration of full-order WBC dynamics models and its application to humanoids.
    Other promising directions include extending the framework with online model adaptation to further improve generalization and precision.


    \section{Acknowledgment}
    \label{sec:acknowledgement} We thank Zijun~Hui, Taerim~Yoon for developing
    the hardware testing software stack. We thank Yu (Antares)~Zhang for helping
    the hardware experiments.

    \bibliographystyle{IEEEtranN}
    \bibliography{references.bib}
\end{document}